\documentclass[lettersize,journal]{IEEEtran}
\usepackage{amsmath,amsfonts}
\usepackage{array}
\usepackage[caption=false,font=normalsize,labelfont=sf,textfont=sf]{subfig}
\usepackage{textcomp}
\usepackage{stfloats}
\usepackage{url}
\usepackage{verbatim}
\usepackage{graphicx}
\usepackage{cite}
\usepackage{xcolor}
\usepackage{multirow}
\usepackage{booktabs}
\usepackage{cancel}
\usepackage[ruled,linesnumbered]{algorithm2e}
\begin{document}

\title{Class Agnostic Instance-level Descriptor for Instance Search}

\author{Qi-Ying Sun, Wan-Lei Zhao*, Hui-Ying Xie, Yi-Bo Miao, Chong-Wah Ngo~\IEEEmembership{Staff,~IEEE,}
\thanks{Wan-Lei Zhao, Hui-Ying Xie, and Yi-Bo Miao are with Computer Science Department, Xiamen University, China. Wan-Lei Zhao is the corresponding author, email: wlzhao@xmu.edu.cn.}
\thanks{Qi-Ying Sun is with Institute of Artificial Intelligence, Xiamen University, China.}
\thanks{Chong-Wah Ngo is with School of Computing and Information Systems, Singapore Management University.}
}

\maketitle

\begin{figure*}[!t]
\centering
\includegraphics[width=\textwidth]{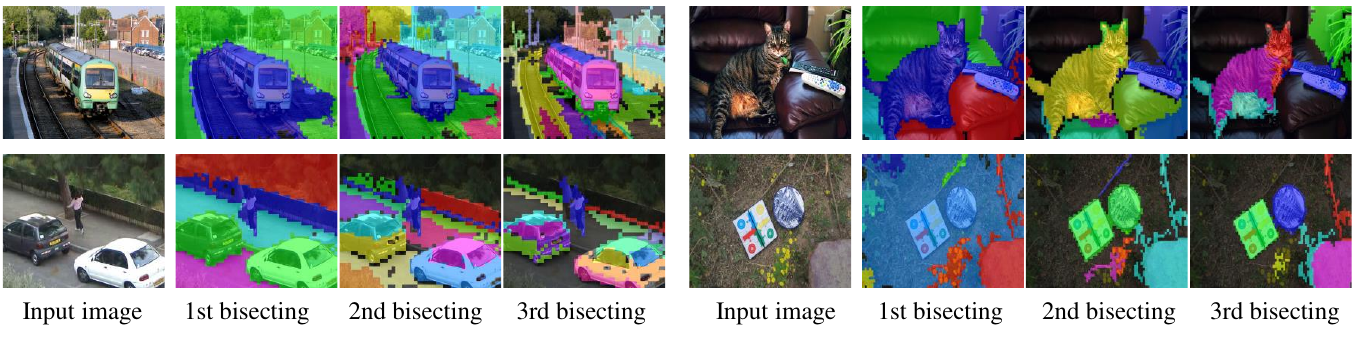}
\caption{Semantic region decomposition in the proposed Class Agnostic Instance-level Descriptor (CLAID). It is a multi-level and multi-granularity instance decomposition, leading to excellent performance on three instance search benchmarks.}
\label{fig:teaser}
\end{figure*}

\begin{abstract}
Despite the great success of the deep features in content-based image retrieval, the visual instance search remains challenging due to the lack of effective instance-level feature representation. Supervised or weakly supervised object detection methods are not the appropriate solutions due to their poor performance on the unknown object categories. In this paper, based on the feature set output from self-supervised ViT, the instance-level region discovery is modeled as detecting the compact feature subsets in a hierarchical fashion. The hierarchical decomposition results in a hierarchy of instance regions. On the one hand, this kind of hierarchical decomposition well addresses the problem of object embedding and occlusions, which are widely observed in real scenarios. On the other hand, the non-leaf nodes and leaf nodes on the hierarchy correspond to the instance regions in different granularities within an image.  Therefore, features in uniform length are produced for these instance regions, which may cover across a dominant image region, an integral of multiple instances, or various individual instances. Such a collection of features allows us to unify the image retrieval, multi-instance search, and instance search into one framework. The empirical studies on three benchmarks show that such an instance-level descriptor remains effective on both the known and unknown object categories. Moreover, the superior performance is achieved on single-instance and multi-instance search, as well as image retrieval tasks.
\end{abstract}


\section{Introduction}
\label{sec:intro}
\IEEEPARstart{V}{isual} instance search is a more challenging search task compared to the conventional content-based image retrieval. It requires the search system to return not only the relevant images in which the target instance is present, but also the localization of the target instance in the relevant images. The localization is usually given as the bounding boxes over the target instance regions. Since the instance only appears as an object in an image sub-region, it could be occluded or undergo various types of transformations. For this reason, features that only describe an image local, such as SIFT~\cite{sift}, SURF~\cite{surf}, and DELF~\cite{delf} are expected. However, although these features are able to represent an object local, a single local feature is usually too fine-grained to form a complete description about an object instance or meaningful part of it (\textit{e.g.}, a logo on a hat). Moreover, these features are vulnerable to object deformations and big viewpoint variations. Due to the above reasons, robust instance-level feature representation, which is able to describe a complete instance or a semantic part of an instance, is expected.

However, it is non-trivial to design an instance-level feature that is suitable for the search task. First of all, the feature detector, which is responsible for the instance region localization, should be class-agnostic. Otherwise, the detector only performs well on the known object categories~\cite{dasr,hong2021towards}. For this reason, supervised or weakly supervised object classification/detection/segmentation models are out of consideration~\cite{kuo2012unsupervised,Zhang2023ACS}. Moreover, the localization of the instance region or its sub-region of various categories is expected. On the one hand, it ensures the target instance returned to the query are precisely localized in the reference images. On the other hand, it enables the instance feature to be derived from a precise instance region, which in turn guarantees the search quality.

Due to the aforementioned challenges, an effective instance-level feature is still slow to occur. In~\cite{cycle-Self}, a cycle self-training framework is proposed for instance search, in which the instance search is addressed by an object tracking framework. The instance region is localized by the correlation operation between the query instance and the candidate image as the GlobalTrack~\cite{aaai20:globaltrack}. Although good performance is achieved on several benchmarks, such a framework is hardly scalable to large-scale search tasks, where the scale of candidate images could be up to a million or even a billion. In contrast to the self-training framework, DASR (Deeply Activated Salient Region)~\cite{dasr} is able to extract semantically significant regions from the candidate images. The salient regions are obtained by back-propagating peaks on the class activation map of an image. Encouraging search performance is reported on two benchmarks. Unfortunately, the instance localization accuracy turns out to be low, in particular for the unknown object categories, as the localization in DASR is built upon a pre-trained classification ConvNet.

Recently, the research in computer vision has been marked by the self-supervised ViT model~\cite{DINO}. The distinctive token-wise feature can be built after the self-distillation. These features are semantically meaningful. They could emerge as the semantic segmentation of an input image. Since the ViT is trained without any labels, it becomes the cornerstone for the unsupervised object/semantic segmentation. With the support of self-supervised ViT, each token on the output layer, which also corresponds to a fixed-size patch on the input image, is modeled as a node in an affinity graph. The unsupervised segmentation task is therefore treated as a graph partition problem in~\cite{cutclr,deepspectral,tokencut}. The decomposed subgraphs correspond to a number of subsets of patches on the input image. Each group of patches constitutes an object region. In particular, TokenCut in~\cite{tokencut} treats the segmentation as a bipartite graph partition problem. One sub-graph is treated as the foreground object, and another is treated as the background. While CutLER in~\cite{cutclr} cuts the graph in a chain. Without referring to any other images or labels, significantly better segmentation accuracy than the previous methods is achieved by CutLER and TokenCut.

In this paper, the instance-level regions are detected by a hierarchical decomposition over the token-wise features from self-supervised ViT. The decomposition is carried out by repeated bisecting clustering as shown in Figure~\ref{fig:teaser}. The repeated bisecting decomposition continues on the resulting sub-regions until the carefully designed termination condition is reached. This process essentially segments one image repeatedly into semantically salient regions. Different from CutLER and TokenCut, we do not simply treat a segmented sub-region as background or a foreground object. As long as it is sufficiently salient, a feature is produced for the decomposed region. With such whole-to-part decomposition, one image is represented by a collection of features that are able to describe the image regions in multiple granularities. On the one hand, such decomposition well addresses the problem of object occlusion and object embedding, which are widely observed in real scenarios. On the other hand, it also builds the feature representation to support image retrieval, multi-instance search, and single instance search. In summary, the contributions of this paper are threefold. 
\begin{itemize}
	\item {We model the problem of instance-level feature detection as a hierarchical decomposition on the patch-level feature set of an image. The decomposition is addressed by stochastic optimization. This leads to an efficient and relatively exhaustive description of the instances/sub-instances latent in an image.}
	\item {The ``terminate condition'' and ``dummy node'' are proposed to support the hierarchical decomposition. The former allows the decomposition to adapt to instances in different sizes and shapes and avoids a sensitive parameter, namely the depth of hierarchical decomposition. The latter helps to filter out many invalid features.}
	\item {The hierarchical decomposition leads to a collection of feature representations for the latent instances in different granularities. They may represent a region large enough to support the similarity judgment between images. They may also represent a region composed of multiple instances or a single instance region.  Therefore, our method allows the image search and instance search to be unified under the same retrieval framework.}
\end{itemize}

To the best of our knowledge, this is the first feature that is able to fulfill three search tasks in the literature. We have made the codes of our work publicly available\footnote{\url{https://github.com/wlzhao22/CLAID}}.
\section{Related Work}
\label{sec:relate}
\subsection{Instance Search}
Before the emergence of deep ConvNet, instance search was addressed as a sub-image retrieval task~\cite{Awad17}. Image local features such as SIFT~\cite{sift} and SURF~\cite{surf} were widely adopted. Essentially, the search seeks the correspondence of local features between the query instance and the candidate images. To speed up the match, feature aggregation methods such as BoVW~\cite{Sivic03} and VLAD~\cite{jegou2011aggregating,chadha2017voronoi} are introduced. Although much better efficiency is achieved, the matching of local features has been degraded to the measurement of similarity between the query instance and the whole candidate image. In addition, due to the sensitivity of local features to the object deformation and 3D rotation, this kind of search paradigm only works on rigid objects.

Due to the wide success of neural networks in many tasks, deep features have been introduced to the instance search task~\cite{r-mac, crow, class-weighted, blcf, RegionalAttention, zhang2023learning,bai2021unsupervised} in the recent decade. Usually, an image is represented by a deep global feature, which is pooled from a certain conv-layer. During the pooling, the higher weights are assigned to the potential instance regions. The feature turns out to be robust to various image transformations, whereas it lacks distinctiveness. In addition, methods of this category are unable to localize the instance from the relevant images.

In the recent studies~\cite{LocalSimilarity, cycle-Self}, the similarity between the query instance and the candidate images is learned via a ConvNet. The correlation is performed between the query instance and each candidate image. High similarity is returned when the target instance appears in the image. Superior performance is reported on several benchmarks. However, the instance search is essentially performed via a deep convolution between the query and all the candidate images. It is hardly scalable to a large scale due to the high time costs. 

Instead of borrowing features designed for other tasks, several attempts~\cite{DeepVision,dasr} have been made to design instance-level features for the search task. In general, there are two major steps involved in the methods, namely instance localization and feature extraction from the detected instance regions. Method in~\cite{DeepVision} is built upon Faster-RCNN~\cite{FasterRCNN}, which limits its capability in detecting objects from unknown categories. In contrast, DASR~\cite{dasr} relies only on a pre-trained image classification network. It shows much stable performance over~\cite{DeepVision} as it remains effective on the unknown object categories. However, the localization is imprecise as the detection fully relies on the class activation map. Although encouraging performance is achieved for the method in~\cite{pami23:zhang}, its performance is still unsatisfactory as its region detector relies on weakly supervised learning.

\subsection{Unsupervised Semantic/Object Segmentation}
Owing to the success of self-distilled ViT DINO~\cite{DINO}, the high quality of unsupervised semantic/object segmentation/detection is achievable in the recent studies~\cite{cutclr,deepspectral,tokencut, cvpr24:u2seg, cvpr24:cuvler, cvpr24:sun}. In these works, the segmentation task is fulfilled by decomposing an affinity matrix, which is built based on the patch-level features derived from DINO. In~\cite{deepspectral}, spectral clustering is adopted for the decomposition. The major eigenvectors after the decomposition correspond to the foreground object in the image. In TokenCut~\cite{tokencut} and CutLER~\cite{cutclr}, the affinity matrix is transformed into a graph by filtering out edges with low similarity. Thereafter, NCut~\cite{ncut} is adopted for the graph partition in both methods. The difference is that only one cut is applied to the graph in TokenCut. The image region that corresponds to the sub-graph with high energy is viewed as an object. While the other is treated as the background. CutLER employs a method called MaskCut~\cite{cutclr}, in which NCut~\cite{ncut} is applied three times. In each round, the resulting background is further divided into a foreground object and a background for the next iteration. It yields considerably higher segmentation accuracy than TokenCut. The coarse masks produced by this decomposition are subsequently refined with a pairwise CRF~\cite{deepspectral}. MaskCut~\cite{cutclr} and NCut~\cite{ncut} are adopted in U2Seg and VoteCut, respectively, to cut an image into semantic regions. In VoteCut~\cite{cvpr24:sun}, an image is cut into several regions based on the feature set produced by six ViT backbones, respectively. The final segmentation is fused by pixel-wise voting from the clustered regions. Compared to CutLER, a more precise semantic segmentation of the major objects in one image can be produced.

Intuitively, methods in unsupervised semantic/object segmentation can be directly borrowed as the detector for the instance search. However, these two tasks are still essentially different. First of all, one cannot assume that the submitted instance query is a complete object. The visual query could be a meaningful sub-region of an object, \textit{i.e.} a logo on the coat or a cap of the bottle. Or on another extreme, the query is comprised by several visual objects, which is known as multiple-instance search. Theoretically speaking, any semantic part of an image could be cropped out as the query. The instance-level feature detector should detect these semantic regions from one image as much as possible. 

Different from TokenCut~\cite{tokencut} and CutLER~\cite{cutclr}, a novel hierarchical decomposition procedure is designed in our paper. Given that an image is represented by a collection of patch features output from a ViT layer, one feature corresponds to a patch on the ViT input layer. A bisecting procedure is proposed to partition the feature set repeatedly into a bunch of subsets. Each subset corresponds to a semantic region in the image. Instead of simply fixing the number of layers of such bisecting, a condition is designed to judge whether the bisecting should terminate at a certain subset. As a result, similar to the conventional local feature, the collection of such features makes up a full description for the whole image. Whereas, unlike the conventional local feature, the descriptions are instance-level and semantically meaningful. More importantly, in contrast to conventional local features, typically only around \textit{30} descriptors are extracted from one image, which largely relieves the burden of afterwards indexing and nearest-neighbor search.

\section{Instance-Level Feature Detector}
\label{sec:detect}
\subsection{Feature Detection by Hierarchical Decomposition}
In ViT~\cite{vit}, an image is partitioned into small patches of equal size like a jigsaw puzzle, and an instance inside the image is composed of dozens of such small patches. When feeding into the backbone, one patch is represented as a feature vector. Given $X$ is the set of features produced by the backbone, an instance inside the image corresponds to a subset of features. According to ~\cite{DINO}, the pairwise feature distances within the subset are expected to be small, while their distances to the features outside the subset should be relatively larger. Therefore, detecting semantically salient regions could be modeled as a problem of clustering that minimizes the intra-cluster distances. Given there are \textit{k} clusters $\{S_1,\cdots,S_k\}$, our objective is
\begin{equation}
	\mbox{Min. } \sum_{r=1}^k\sum_{i,j \in S_r \& i < j}\parallel x_i - x_j \parallel^2.
	\label{eqn:ksumis}
\end{equation}
Although the objective looks similar to that of traditional \textit{k}-means at first sight, it is essentially different from traditional \textit{k}-means and most of its variants. The objective minimizes the summation of pairwise L2 distance within each cluster. Intuitively, we have to try out all the possible combinations of the feature vectors across different clusters to seek the optimal solution for Eqn.~(\ref{eqn:ksumis}), which is, however, computationally prohibitive. Moreover, the egg-chicken loop in the traditional \textit{k}-means is no longer applicable as no cluster centroid is defined in Eqn.~(\ref{eqn:ksumis}).

In our paper, the procedure proposed in \textit{k}-sums~\cite{cikm21:ksum} is employed to address this optimization problem. Given $x_i \in S_w$, the aggregated distance between sample $x_i$ and the rest of feature points in $S_w$ is defined as

\begin{equation}
	d(x_i, S_w)=\sum_{x_j \in S_w}\parallel x_i - x_j \parallel^2.
	\label{eqn:x2Sr}
\end{equation}
Eqn.~(\ref{eqn:x2Sr}) can be further simplified into following form
\begin{equation}
	d(x_i, S_w)=n_w{\cdot}x_i^T{\cdot}x_i-2{\cdot}x_i^T{\cdot}D_w+\sum_{x_j \in S_w}{x_j^T{\cdot}x_j},
	\label{eqn:x2Srv2}
\end{equation}
where $D_w=\sum_{x_j \in S_w}x_j$ and $n_w=|S_w|$. The first term and the last term in Eqn.~(\ref{eqn:x2Srv2}) are both the L2-norm of the feature vectors. The second term is the inner-product between feature $x_i$ and $D_w$. Since features are L2-normalized in our problem, Eqn.~(\ref{eqn:x2Srv2}) is simplified further as $d(x_i, S_w)=2{\cdot}n_w-2{\cdot}x_i^T{\cdot}D_w$. The minimization on Eqn.~(\ref{eqn:ksumis}) is fulfilled in a stochastic and greedy manner. Each time a feature $x_i \in S_w$ is randomly selected, we want to check whether Eqn.~(\ref{eqn:ksumis}) would decrease assuming that $x_i$ was moved to $S_v$ already,
\begin{equation}
	\label{eqn:x2Sv}
	\begin{aligned}
		d(x_i,S_v)=(n_v&+1){\cdot}x_i^T{\cdot}x_i-2{\cdot}x_i^T{\cdot}(D_v+x_i)\\
		&+\sum_{x_j \in S_v \& j \neq i}{x_j^T{\cdot}x_j} +x_i^T{\cdot}x_i\\
		=n_v{\cdot}&x_i^T{\cdot}x_i-2{\cdot}x_i^T{\cdot}D_v+\sum_{x_j \in S_v \& j \neq i}{x_j^T{\cdot}x_j}.\\
	\end{aligned}
\end{equation}

If $d(x_i,S_v) < d(x_i,S_w)$, such movement would lead to the decrease of objective Eqn.~(\ref{eqn:ksumis}). In particular, we can check over $k-1$ clusters to seek the most greedy movement that Eqn.~(\ref{eqn:ksumis}) decreases the most. In one round, each feature $x_i$ is selected at random to seek the most greedy movement. While if $d(x_i,S_v) \geq d(x_i,S_w)$, the movement will not be undertaken. The greedy optimization usually converges to a local optima~\cite{cikm21:ksum}. Since the step that seeks the best movement is comparable to seeking the closest centroid in traditional \textit{k}-means, the computational complexity of the above optimization is on the same par as the traditional \textit{k}-means. 

\textbf{Repeated Bisecting} The primary goal of clustering is to distinguish different instance regions. However, it is not appropriate to segment the image directly into \textit{k} regions. First of all, the hyperparameter \textit{k} is hard to set. The number of potential instance regions varies from one image to another. Moreover, our perception of visual instances is on multiple scales. For instance, the logo on the bottle is an instance. However, it is only a part of another instance when we view the bottle as an instance. Furthermore, when the bottle is on a desk, the bottle is just a small instance embedded in another bigger instance ``desk''. In order to mimic the human perception of the visual instances, the clustering is carried out hierarchically. Specifically, the feature set is bisected into two each time. For a complete image as input, in each step, we decompose the image or its sub-image into two fine-grained sub-regions. Different levels of decomposition locate visual instances of different scales. A termination condition, which will be detailed later in this section, is designed to decide whether the bisecting should not be carried out further on a subset. In this way, we avoid setting the hyper-parameter \textit{k} explicitly. The number of clusters we finally have is largely related to the number of latent instances in an image.

Given a node on the hierarchy, one more step is undertaken to partition the resulting subsets further into smaller ones based on the spatial connectivity of patches. Therefore, the hierarchy is not a binary tree in the usual case.

In the following, the details are presented about how the hierarchical detection procedure is initialized and what the condition is to terminate the bisecting on a node. It is possible that the segmented regions are the empty background or a small instance region embedded in a big empty background. It is unnecessary to build a feature for such regions. A condition is further designed to identify these nodes on the hierarchy.

\subsection{Decomposition Seed Selection}
\label{sec:seed}
Given a feature set (or a node) $S$, the bisecting clustering starts from two randomly bisected clusters on $S$~\cite{cikm21:ksum}. Due to the random initialization, the clustering result varies from one run to another. To address this issue, a prior is introduced to initialize the bisecting. To facilitate our discussion, affinity matrix A is defined as follows
\begin{equation}
	A_{ij} = \begin{cases}
		1, &  x_i^T{\cdot}x_j>\alpha,\\
		0,& otherwise,\\
	\end{cases}
	\label{eq:affinity1}
\end{equation}
where $x_i, x_j \in S$, and $\alpha$ is empirically set to \textit{0.2}\cite{bmvc21:lost}. Given $x_b \in S$ is the feature holds the lowest degree according to A (given in Eqn.~(\ref{eq:xb})) and $x_w$ (given in Eqn.~(\ref{eq:xw})) is the feature holds the highest degree, they are most likely located on different instances as discovered in~\cite{bmvc21:lost}.

\begin{equation}
	x_b = x_{i_{min}}, \mbox{where~} i_{min} = \underset {i} {\operatorname {arg\,min}} \, \sum_j A_{ij} 
	\label{eq:xb}
\end{equation}

\begin{equation}
	x_w = x_{i_{max}}, \mbox{where~} i_{max} = \underset {i} {\operatorname {arg\,max}} \, \sum_j A_{ij} 
	\label{eq:xw}
\end{equation}

\begin{figure}[t]
	\centering
	\includegraphics[width=\linewidth]{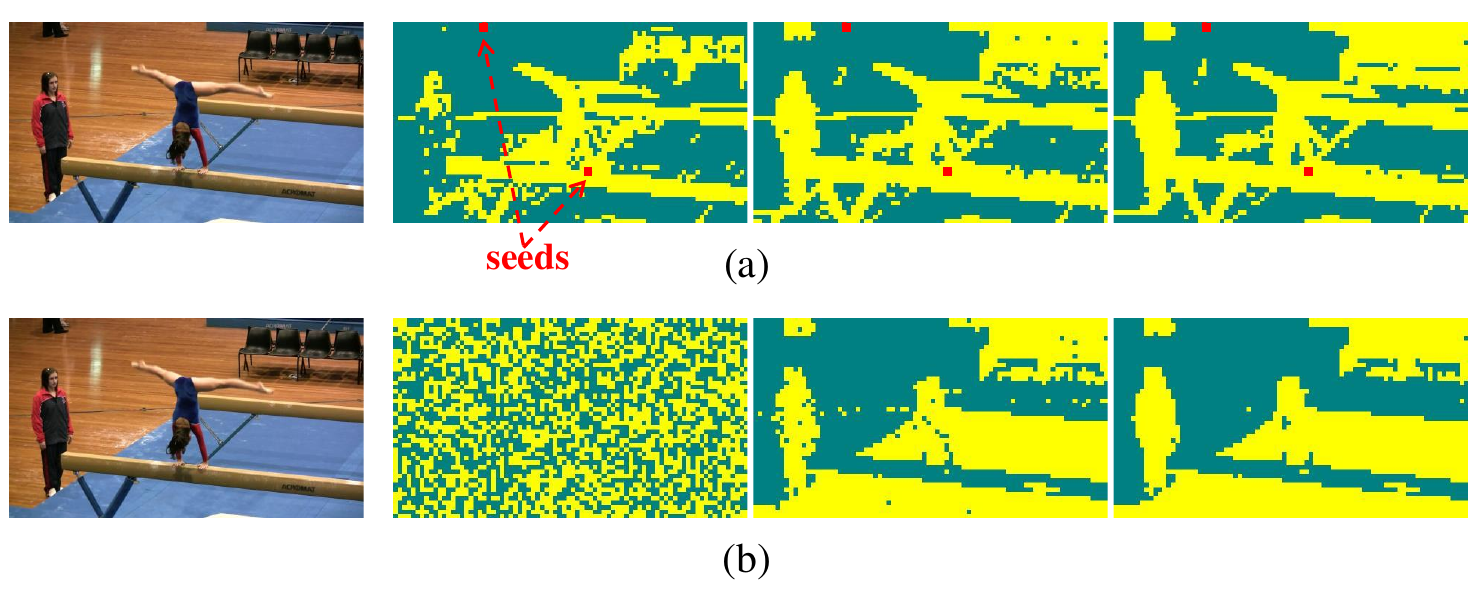}
	\caption{The process diagram showcases different initialization methods for the bisecting clustering. The top row illustrates the clustering with seed selection. The first column shows the original image, followed by the images after the initialization, intermediate clustering result, and the final clustering results. The bottom row shows the clustering under random initialization.}
	\label{fig:process}
\end{figure}

So $x_b$ and $x_w$ are selected as two seeds for the bisecting. The rest of the features in $S$ are assigned to a cluster of $x_b$ or $x_w$ by checking whether they are closer to $x_b$ or $x_w$. The initialization scheme is adopted for the bisecting of all the nodes on the hierarchy. The difference between the clustering with seed selection and clustering with random initialization is illustrated in Figure~\ref{fig:process}.

\subsection{Terminate Condition}
\label{sec:terminate}
\begin{figure}[t]
\begin{center}
	\subfloat[input image]
	{\includegraphics[width=0.8\linewidth]{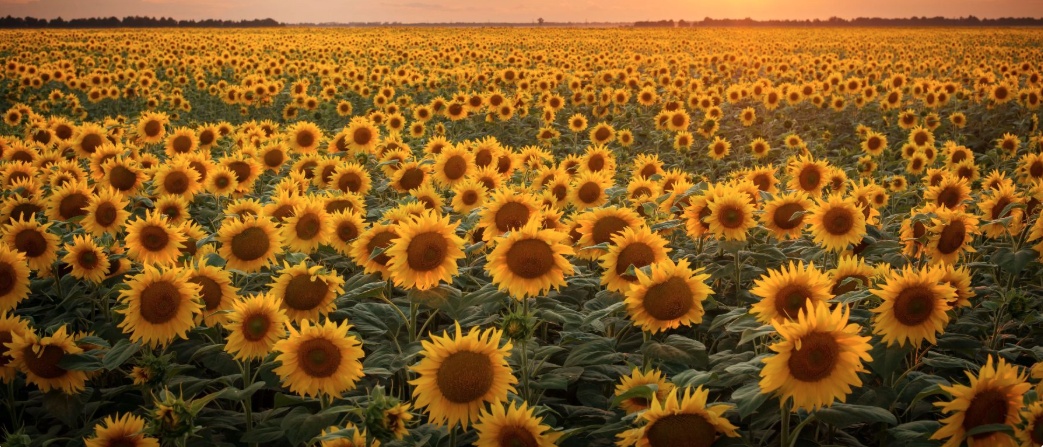}}\\
	\subfloat[instance regions detected by CLAID]
	{\includegraphics[width=0.8\linewidth]{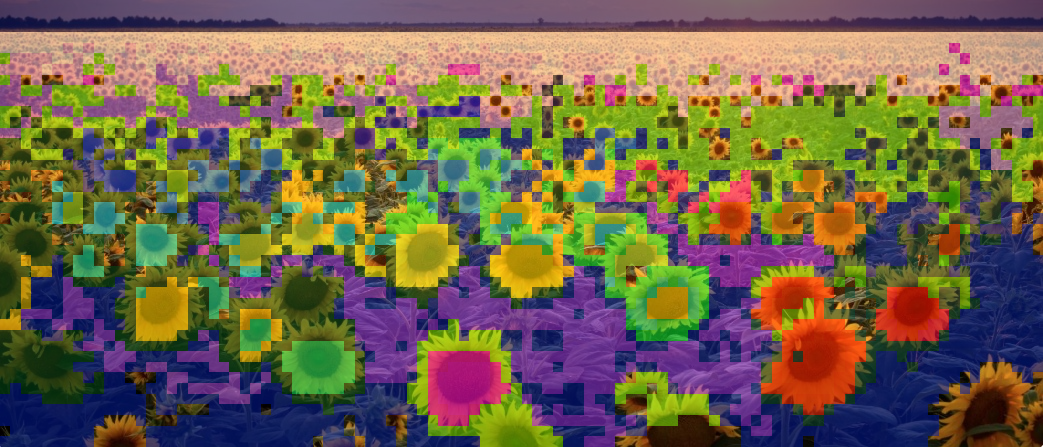}}
\end{center}
\caption{An illustration of detecting dense objects in different sizes and shapes by CLAID. Different instances are covered by different colors. For clarity, the instances correspond to the non-leaf nodes are not shown in figure (b).}
\label{fig:sunflower}
\end{figure}
Given a feature set (or a node) $S$ on the hierarchy, the total number of edges within the feature set $S$ is 
\begin{equation}
	c(S) =\sum_{i,j \in S \& i < j} A_{ij}.
	\label{eq:distance1}
\end{equation}
The average internal connectivity within $S$ is defined as 
\begin{equation}
	\bar{c}(S) = 2\cdot\frac{c(S)}{n\cdot(n-1)},
	\label{eq:connect}
\end{equation}
where $n=|S|$. Apparently, $\bar{c}(S)$ is in the range $[0, 1]$. Since features inside $S$ correspond to a collection of fixed-size patches on the image, the higher $\bar{c}(S)$ is, the more these patches are semantically closer to each other. Compared to the parent node, $\bar{c}{(S)}$ from the child nodes is usually greater. As the decomposition continues, $\bar{c}{(S)}$ of the decomposed node becomes closer to \textit{1.0}. The bisecting over $S$ should not be carried out further when the connectivity within a feature set is sufficiently high. At this stage, the integral of these features may represent an inseparable instance region in the image. To this end, we can control the decomposition granularity by tuning the threshold $\tau_1$ on $\bar{c}(S)$. A larger threshold leads to finer decomposition granularity. In our implementation, when $\bar{c}(S) \geq {\tau_1}$, the bisecting on $S$ is not undertaken further.

Subjected to the granularity of ViT, the patch size in ViT cannot be arbitrarily small. The coverage of a ViT patch could be bigger than the size of a small instance. The fine-grained feature description of the small object becomes impossible. Therefore, the localization of such a small instance cannot be achieved by our method. Instead, a collection of dense small instances will be detected as a big virtual instance, namely an instance containing a group of dense small instances. This may not be meaningful for the semantic segmentation. Nevertheless, this is reasonable in the context of instance search. Since these small instances lose their individual details as objects, it is more realistic to treat them as a whole to be a subject in the image/instance search.

Figure~\ref{fig:sunflower} shows a typical example where dense objects are displayed in different sizes and shapes. As shown in the figure, our method is able to adapt to objects in different shapes and sizes. The dense small sunflowers in the distance are detected as a big instance (covered by yellow).

\subsection{Condition on Dummy Nodes}
\label{sec:dummy}
\begin{figure}[t]
\vspace{-0.12in}
	\centering
	\subfloat[input image]{%
        \includegraphics[width=0.48\linewidth]{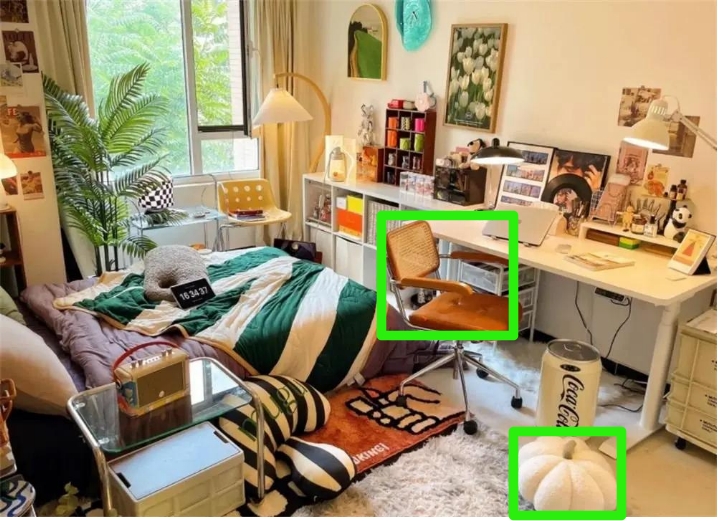}
    }
    \hfill
    \subfloat[high energy regions]{%
        \includegraphics[width=0.48\linewidth]{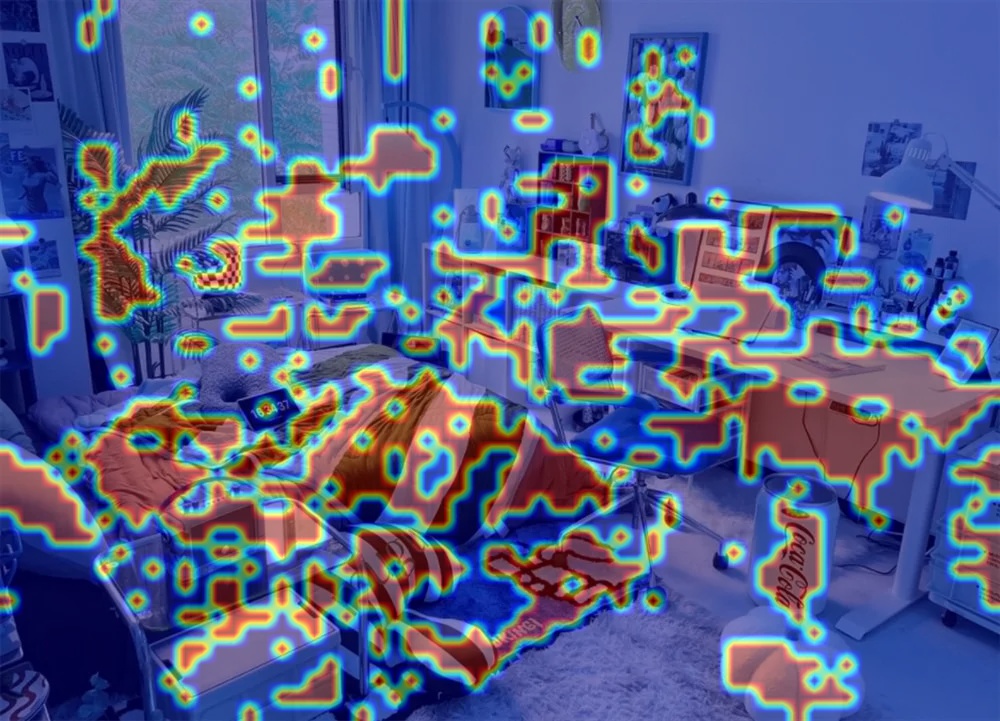}
    }\\
    \subfloat[dummy node (in blue)]{%
        \includegraphics[width=0.48\linewidth]{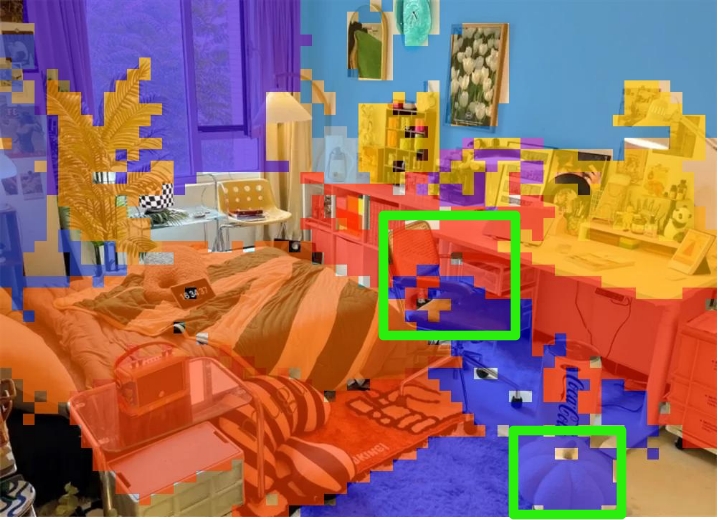}
    }
    \hfill
    \subfloat[latent instances]{%
        \includegraphics[width=0.48\linewidth]{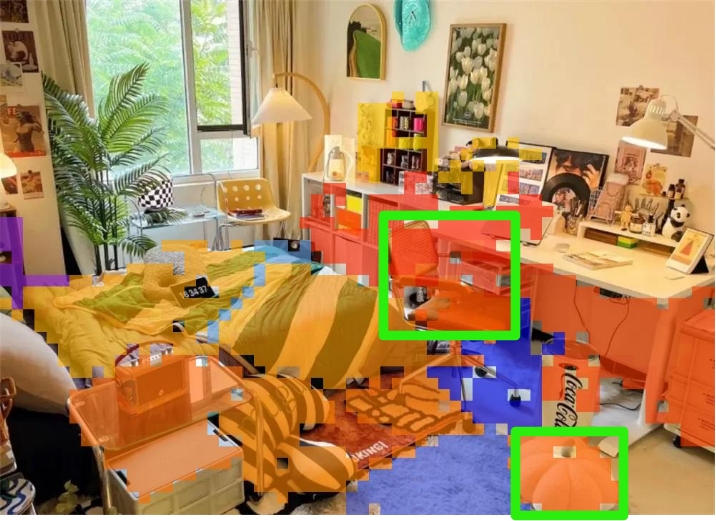}
    }
    \caption{The illustration of visual instances under the ``dummy node'' in a real scenario. A region (in blue) in Figure (c) is detected as ``dummy node'' based on Eqn.~(\ref{eq:salience}) after the \textit{3rd} bisecting. Two latent instances are detected when we further decompose the ``dummy node'' at the \textit{5th} bisecting. Figure (b) visualizes the high energy regions, i.e., $H$ (in red color) in the image.}
    \label{fig:dummynode}
\end{figure}

Given a feature set (or a node) $S$ on the hierarchy, it does not necessarily correspond to a semantically salient region. It could be the mixture of objects and the background, or simply an empty background. We should not build a feature for such a region. Firstly, it is unnecessary to build a feature if the region is empty. Secondly, for the region that an instance is embedded in a big, empty background, the feature representation could be mixed with noise due to the dominant background. It would be more appropriate to build the feature when the empty background has been decomposed. We view such a region as a ``dummy node'' in the decomposition hierarchy. Inspired by~\cite{l1norm}, ``dummy node'' is judged by the strength of feature energy. In our method, we define $H=\arg\max\nolimits_{\forall x_i \in X}^{\theta} \left| x_i \right|^1$ as the set of features that exhibit high energy within the dataset $X$. Here, we set $\theta$ to the top-$30\%$ based on the relevant experimental results presented in \cite{l1norm}.
We measure the overlapping ratio between $S$ and $H$
\begin{equation}
	\xi(S) =\frac{\left| H \cap S \right|}{ \left| S \right|}.
	\label{eq:salience}
\end{equation}
When $\xi(S)$ is low, it basically indicates the region is over-dominated by semantically insignificant features.In our implementation, a node on the bisecting hierarchy is viewed as ``dummy'' if $\xi(S) \leq \tau_2$, where $\tau_2$ is empirically set. No feature will be built from it when a detected region is viewed as a ``dummy node''. However, this condition does not prevent the feature set $S$ from further bisecting. Since $\xi(S)$ is a relative ratio, $\xi(S)$ may become higher when the bisecting reaches a more fine-grained level. Therefore, the direct/indirect subset of $S$ may become a salient region.

Figure~\ref{fig:dummynode} shows an example where ``dummy node'' is detected. After the \textit{3rd} bisecting, the region in blue (Figure~\ref{fig:dummynode}(c)) is detected as ``dummy node'' for its relatively low intersection ratio with $H$. We do not simply ignore such a region. Instead, further bisecting is applied to the region. After another two times of bisecting on this node, two latent instances are successfully detected because their relative intersection ratio (i.e., $\xi(S)$) becomes much higher this time.

\begin{figure*}[t]
	\centering
	\includegraphics[width=\linewidth]{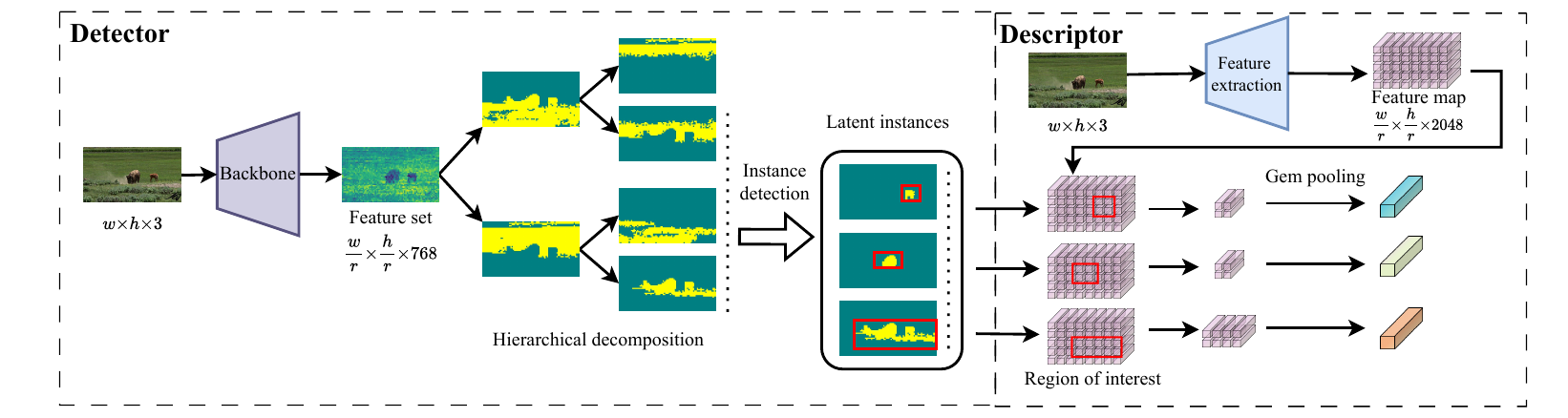}
	\caption{The framework for building instance-level features for instance search. There are two major components: an instance-level region detector and a feature descriptor. Given an image, a set of patch-level features is produced by the self-supervised backbone. The hierarchical decomposition is applied to the feature set. Each node on the hierarchy corresponds to a potential instance region or sub-region. The patch-level masks can be produced for the valid nodes on the hierarchy. The feature for each region is pooled from another network. Given an incoming query, a feature is extracted with the same backbone. }
	\label{fig:frame}
\end{figure*}

\begin{algorithm}[t]
	\caption{Instance-Level Feature Detector}
	\label{alg:algorithm1}
	\KwIn{Input image $I$}
	\KwOut{Set of regions $\mathcal{R}$.}  
	\BlankLine
	Initialize $X = backbone(I)$ \;
	Initialize $\mathcal{L} \gets X$ and $\mathcal{R} \gets \emptyset $\;
	\While{$\mathcal{L} \neq \emptyset$}
	{Pop $node$ from $\mathcal{L}$\;
		\If{$\xi(node) > \tau_2$ and $node \neq root$}{$\mathcal{R}$.append($node$);}
		\If{$\bar{c}(node) < \tau_1$}{$S_b, S_w \gets$ \textit{k}-sum($node$)\;
			$\mathcal{L} \gets \mathcal{L} \cup \text{getObject}(S_b) \cup \text{getObject}(S_w)$\;
			}
	}
\end{algorithm}

The entire process of instance-level detection is summarized in Algorithm~\ref{alg:algorithm1}. The hierarchical decomposition starts from the feature set $X$. The bisecting is repeatedly applied to $X$ and its subsets. The resulting subsets $S_b$ and $S_w$ do not necessarily correspond to two instance regions. Several instances from the same object category are likely partitioned into one subset. As a result, one more step ``$getObject(\cdot)$'' (Algorithm~\ref{alg:algorithm1} \textit{Line -- 10}) is undertaken to partition subsets $S_b$ and $S_w$ respectively into spatially connected smaller subsets. ``$getObject(\cdot)$'' simply divides a subset into spatially connected regions by the eight-connectivity criterion, where tiny regions are ignored. The detection process results in a list of coherent candidate instances and their masks. Two subsidiary conditions are involved in the algorithm. Specifically, if a node is dummy according to $\xi(node)$, it will not be joined into the region list. $\bar{c}(node)$ determines whether a node should be bisected further.

\textbf{Discussion} Compared to CutLER~\cite{cutclr} and TokenCut~\cite{tokencut}, we do not simply treat the two regions after the decomposition as either foreground or background. Instead, we believe that there are latent instances in both regions. Secondly, instead of setting a threshold on the number of decompositions explicitly~\cite{tokencut,cutclr}, we introduce the threshold $\tau_1$ on the average internal connectivity. It basically reflects the internal semantic similarities of image patches within a region. When the patches within a region show the same or similar semantics (reflected as a high $\bar{c}(S)$ value), it is believed to be an inseparable instance region and no need to be decomposed further. Moreover, compared to NCut~\cite{ncut} adopted by CutLER and TokenCut, the bisecting by \textit{k}-sums~\cite{cikm21:ksum} employed in our method is much efficient. As will be revealed in the experiments, it also adapts well to the multi-scale semantic structure of an image. Additionally, the introduction of ``dummy node'', on the one hand, largely reduces the number of invalid regions that are dominated by semantically insignificant patches. On the other hand, it also allows us to discover small instances that are embedded in the big background.\section{Feature Descriptor}
\label{sec:desc}
The output from the repeated bisecting on the feature set $X$ is a hierarchy. Each node on the hierarchy is a subset of $X$ (except for the root node), which corresponds to a potential salient region. In this section, we show how the features are extracted for these detected salient regions, except for the dummy nodes on this hierarchy (discussed in Section~\ref{sec:dummy}).

Since the features are on the patch level, there is a one-by-one correspondence between the feature and a patch on the ViT input layer. The segmented subset on the hierarchy appears as a bunch of connected patches on the image. The detected region is given as a region with a patch-level mask. Unlike unsupervised object detection or semantic segmentation~\cite{cutclr, bmvc21:lost}, fine-grained localization is not necessary for instance search. Therefore, the post-processing steps such as up-sampling and boundary modification of location results, such as CRF~\cite{crf} are not adopted in our work.

Given the location of the salient region is available, the feature extraction is as easy as performing RoI-pooling on a feature map from a network layer. To this end, we have multiple options. We can use the CLS token or pooled output of ViT models as feature descriptor, based on which our salient regions are detected. Alternatively, the pre-trained ConvNet, e.g., VGG~\cite{vgg} or ResNet~\cite{resnet} can be used to derive the feature. Moreover, instead of feeding the entire image into the network and pooling based on the region box, the detected region can be cropped out and fed into the network as one image to produce the feature. However, this way is computationally inefficient. For this reason, we choose to feed the image into the pre-trained ResNet and extract the instance-level feature via RoI-pooling.

The complete pipeline of our hierarchical region detection and feature extraction is shown in Figure~\ref{fig:frame}. At the first stage, the potential instance regions or sub-regions are detected via a top-down hierarchical decomposition over the feature set produced by the backbone. Given that the detection is an unsupervised procedure and is built upon class-agnostic self-supervised ViT, and the feature is extracted from a ResNet. Since the detector is built upon class-agnostic self-supervised ViT, the produced feature becomes a class-agnostic descriptor in the real sense. For this reason, our feature is called \textit{CLass Agnostic Instance-level Descriptor} (CLAID) from now on. In contrast to traditional local features, only \textit{30} instance-level regions per image are extracted. For an instance query, the cropped query region is fed into the ``Descriptor'' block directly. Only one feature vector is produced for a query. 

\begin{table*}[t]
        \caption{Search performance (mAP) on \textit{Instance-160}, \textit{Instance-335}, and \textit{INSTRE}. $^\ddag$ digits are collected from the referred paper}
	\centering
	\fontsize{8}{7}\selectfont 
	\begin{tabular}{lcccccccccc}
		\toprule
		\multirow{2}{*}{Method}&
          Descriptor &	\multicolumn{3}{c}{Instance-160} & \multicolumn{3}{c}{Instance-335} & \multicolumn{3}{c}{INSTRE} \\
		\cmidrule(lr){3-5}
		\cmidrule(lr){6-8}
		\cmidrule(lr){9-11} 
		& backbone & mAP-50 & mAP-100 & mAP-all & mAP-50 & mAP-100 & mAP-all & mAP-50 & mAP-100 & mAP-all \\
		\cmidrule(lr){1-11}
		R-MAC~\cite{r-mac} & VGG-16 &0.302 & 0.281 & 0.330 & 0.417 & 0.354 & 0.381 & 0.741 & 0.626 & 0.523 \\
		CAM-weight~\cite{class-weighted} & VGG-16 & 0.337 & 0.310 & 0.358 & 0.404 & 0.336 & 0.347 & 0.521 & 0.388 & 0.320 \\
		BLCF~\cite{blcf} & VGG-16 & 0.641 & 0.617 & 0.651 & 0.506 & 0.460 & 0.483 & 0.657 & 0.578 & 0.636 \\
		BLCF-SalGAN~\cite{blcf} & VGG-16 & 0.633 & 0.604 & 0.638 & 0.490 & 0.442 & 0.470 & 0.821 & 0.751 & 0.698 \\
		DASR*~\cite{dasr} & ResNet-50 & 0.752 & 0.720 & 0.771 & 0.783 & 0.705 & 0.724 & 0.870 & 0.779 & 0.692 \\
		CutLER~\cite{cutclr}& ResNet-50 &0.675 &0.635 &0.683 &0.712 &0.630 &0.641 &0.870 &0.799 & 0.727\\
        DUODIS~\cite{pami23:zhang}$^\ddag$& VGG-16 &- &- &- &- &- &- &- &- & 0.667\\
        DUODIS~\cite{pami23:zhang}$^\ddag$& ResNet-101 &- &- &- &- &- &- &- &- & 0.752\\
        U2Seg~\cite{cvpr24:u2seg} & ResNet-50 & 0.708& 0.667 & 0.729 & 0.723 & 0.643 & 0.669 & 0.869 & 0.790 & 0.714\\
        VoteCut~\cite{cvpr24:cuvler} & ResNet-50 & 0.720 & 0.677 & 0.725 & 0.749 & 0.664 & 0.682 & 0.884 & 0.817 & 0.761\\
        Orth~\cite{cvpr24:sun} & ResNet-50 & 0.768 & 0.725 & 0.776 & 0.783 & 0.700 & 0.716 & 0.879 & 0.809 & 0.756\\
		\midrule
		CLAID~(NCut)& ResNet-50 &0.771 &0.732&0.780 &0.785 &0.703 &0.722 &0.889 &0.822 & 0.765\\
        CLAID& VGG-16 &0.686 &0.635 &0.681 &0.674 &0.581 &0.591 &0.851 &0.755 & 0.667\\
		CLAID& ResNet-50 &0.781 &0.743 &0.791 &0.789 &0.710 &0.725 &\textbf{0.894} &\textbf{0.827} &\textbf{0.775}\\ 
        CLAID& ResNet-101 &\textbf{0.786} &\textbf{0.745} &\textbf{0.797} &\textbf{0.793} &\textbf{0.715} &\textbf{0.735} &0.893 &0.825 &0.773\\
		\bottomrule
	\end{tabular}
	\label{tab:3search}
\end{table*}
\section{Experiment}
\label{sec:exp}
\subsection{Experiment Setup}
\begin{table}

\begin{center}
\caption{Summary over datasets used in the performance evaluation}
\begin{tabular}{llrr} \toprule
Task & Dataset & \#Queries & \#Images \\ \hline
\multirow{4}{*}{Instance Search} & Instance-160 & 160 & 11,885  \\
& Instance-335 & 335 & 40,914 \\ 
& INSTRE & 1,250 & 27,293 \\ 
& INSTRE-M & 250 & 27,293 \\ \hline
\multirow{3}{*}{Image Retrieval} & Holidays & 500 & 1,491 \\
& Oxford5K & 55 & 5,062 \\
& Paris6K  & 55 & 6,412 \\ \bottomrule
\end{tabular}
\label{tab:datasets}
\end{center}
\end{table}

The effectiveness of the proposed method CLAID is evaluated on both single instance search, multi-instance search, as well as image retrieval tasks. Three popular benchmarks in instance search are adopted. They are \textit{Instance-160}, \textit{Instance-335}\footnote{https://drive.google.com/file/d/1ZxVUO3uInFqiFzXfuokQxalwMl1fEDgN\\/view?usp=drive\_link} and \textit{INSTRE}~\cite{instre}. \textit{Instance-335} is an augmented dataset of \textit{Instance-160} with data collected from the GOT-10K~\cite{GOT-10K}, Youtube BoundingBoxes~\cite{youtube}, and LaSOT~\cite{lasot} datasets. In \textit{INSTRE}, there are \textit{250} queries containing multiple instances. In order to verify the performance of CLAID in the multiple instance search task, the performance on this subset of queries (given as \textit{INSTRE-M}) is further studied. In the image retrieval task, Holidays, Oxford5K, and Paris6K are adopted. The major information about these datasets are summarized in Tab.~\ref{tab:datasets}. The search quality is measured by mAP. Specifically, mAP-50, mAP-100, and mAP-all are reported for instance search. mAP-all is reported for the image retrieval task.

In addition, we also report the instance localization accuracy on three datasets, which is measured by the commonly used Intersection Over Union (IoU). In order to study the exhaustiveness of our detector in discovering the latent instance regions in one image, we test it as an unsupervised object detection method. We study the recall of the correct localizations on \textit{COCO} 2017 validation dataset with stuff and instance annotations~\cite{mscoco} in comparison to state-of-the-art methods CutLER~\cite{cutclr}, TokenCut~\cite{tokencut}, Deep Spectral~\cite{deepspectral}, and VoteCut~\cite{cvpr24:cuvler}

Theoretically speaking, a detector can be built upon any ViT network. Due to its outstanding performance, ViT-Base (ViT-B/8) trained by DINO~\cite{DINO} is selected as the default backbone. For the feature extraction backbone, there could be several options as well. ResNet-50~\cite{resnet} is selected in our standard configuration following the same setting in ~\cite{dasr} where the feature is derived from Block4.0. Moreover, the search performance based on the other backbones such as ResNet-101~\cite{resnet} and VGG-16~\cite{vgg} is presented to validate the effectiveness of our detector. Both the detector and descriptor were implemented using PyTorch. The experiments were conducted on an Nvidia RTX 4090 GPU.

\begin{figure*}[t]
	\centering
	\includegraphics[width=\linewidth]{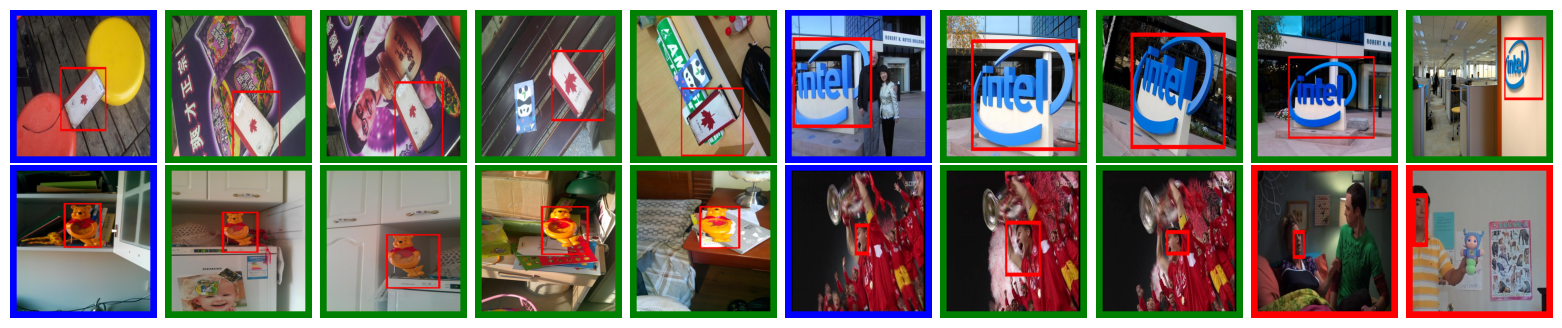}
	\caption{This display showcases the results of instance search by CLAID. The instance queries are from images enclosed in a blue border. Each query is followed by  four retrieved images from the top-1, top-5, top-10, and top-50 outcomes for that query. True-positives are colored in green, while false-positives are in red. CLAID performs well in the majority of queries. Nevertheless, its performance degrades on blurry images or semantically specific queries (e.g., a specific human face).}
	\label{fig:visualres}
\end{figure*}

\subsection{Instance Search Performance}
In the first experiment, the effectiveness of our method is studied in comparison to several representative methods in instance search. Among these methods, R-MAC~\cite{r-mac}, CAM-weight~\cite{cam}, BLCF~\cite{blcf}, and BLCF-SalGAN~\cite{blcf} are deep global features that collapse all the features of an image into one vector for retrieval. In contrast, methods like DASR~\cite{dasr} and ours extract the instance-level feature from the candidate images and are able to localize the instance from the candidate images. In addition, we also treat the unsupervised object detectors CutLER~\cite{cutclr},  U2Seg~\cite{cvpr24:u2seg},  VoteCut~\cite{cvpr24:cuvler}, and Orth~\cite{cvpr24:sun} as the instance region detectors. The same feature as ours is built for the regions detected by CutLER, U2Seg, VoteCut, and Orth, respectively. Moreover, we also report the results when the bisecting clustering in CLAID is replaced by NCut~\cite{ncut}, while keeping other configurations unchanged. The instance search performance is shown in Table~\ref{tab:3search}.

As seen from the table, global features show poor performance on the instance search task. Object level features such as DASR, CutLER, U2Seg, VoteCut, Orth, and CLAID demonstrate considerably better performance. CLAID outperforms the rest of the methods on three benchmarks. Although methods such as CutLER, U2Seg, VoteCut, and Orth are also able to build class-agnostic instance-level features, their performance on instance search is inferior to CLAID. The relatively poor performance is mainly because they only detect the main objects in one image. Typically, the cut in a chain in CutLER causes the loss of potential instance regions. Similar comments apply to U2Seg and VoteCut. Compared to CLAID (NCut), CLAID performs considerably better. According to our observation, the results from bisecting clustering better reflect the multi-scale semantic structure than the partitions produced by NCut. In addition, we also test CLAID with different backbones at its feature extraction stage. As shown in the table, our method shows considerably better performance when it is supported by the same backbone as the competing methods (e.g., DASR$^*$ and CutLER). 

Figure~\ref{fig:visualres} further showcases the instance search results from CLAID. More instance search results can be found from the Appendix. Meanwhile, the performance on other pre-trained ViT networks such as DINOv2~\cite{dinov2}, SigLIP~\cite{Zhai2023SigLIP},  and ViT pretrained by ~\cite{vit} is also reported in the Appendix.

In order to study the behavior of our method on a multi-instance search task, the performance on \textit{250} queries from \textit{INSTRE} is particularly evaluated. For these queries, there is more than one object present in a query. As shown in Table~\ref{tab:multi}, most of the methods show relatively better performance on this task than that of single instance search (reported in Table~\ref{tab:3search}). The search quality from all global features is considerably better. This is mainly because the large coverage of the instance query over the image makes it close to a conventional content-based image search task. The mixture of features from different instances becomes helpful in this context. Our method, along with recent class-agnostic detectors such as CutLER, U2Seg, VoteCut, and Orth, shows the best performance. All these methods are able to detect the major instance regions in one image, which are usually an integral of multiple instances.
\begin{table}
    \caption{Performance (mAP) on multi-instance search task}
	\centering
		\fontsize{8}{7}\selectfont 
	\begin{tabular}{lccc}
		\toprule
		Methods & mAP-50&mAP-100&mAP-all\\
		\midrule
		R-MAC~\cite{r-mac} & 0.824&0.742&0.465 \\
		CAM-weight~\cite{cam} & 0.593&0.464&0.267  \\
		BLCF-SalGAN~\cite{blcf}  & 0.938&0.914&0.750 \\
		DASR*~\cite{dasr} & 0.962&0.916&0.692 \\
		CutLER~\cite{cutclr} & 0.964&0.942&0.768 \\
		U2Seg~\cite{cvpr24:u2seg}& 0.957& 0.928& 0.729\\
		VoteCut~\cite{cvpr24:cuvler}& \textbf{0.968}& \textbf{0.949}& \textbf{0.783}\\
		Orth~\cite{cvpr24:sun}& 0.967& 0.946& 0.773\\
		CLAID &0.964& 0.943& 0.774\\
		\bottomrule
	\end{tabular}
	\label{tab:multi}
\end{table}

The localization accuracy of the target instances on the candidate images is reported in Table~\ref{tab:iou}. Since the localization is impossible to fulfill for the global features, only the performance from the level features is reported. The patch-level region masks from our method, CutLER, U2Seg, VoteCut, and Orth are transformed to bounding boxes when we calculate the mIoU score. As shown in Table~\ref{tab:iou}, CLAID shows the secondary localization accuracy to VoteCut, which is the result of voting from several CutLER-like detectors based on six ViT backbones. The high localization accuracy is at the cost of relatively low detection efficiency (as shown in Table~\ref{tab:speed}).

\subsection{The Performance of Feature Detector}
\begin{figure}
	\centering
	\includegraphics[width=0.6\linewidth]{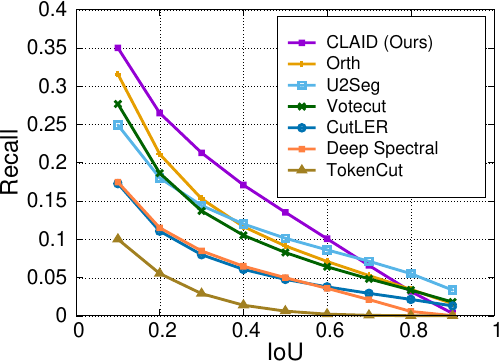}
	\caption{Recall at different IoU thresholds for seven methods on \textit{COCO 20k}.}
	\label{fig:exhaust}
\end{figure}
Since the feature detector is the critical component in our pipeline, its effectiveness is further studied from two perspectives, namely the detection speed and the exhaustiveness in detecting latent instance regions from an image. The processing speed is compared to unsupervised object detectors such as Deep Spectral~\cite{deepspectral}, TokenCut~\cite{tokencut}, and CutLER~\cite{cutclr}. All of them are built on self-supervised ViT. For this reason, we only evaluate the time cost of processing one image. The speed efficiency is evaluated on CPU with a single thread. 

The average number of cuts that are carried out on one image, the average time of processing one image, and the time cost of one cut are shown in Table~\ref{tab:speed}. In TokenCut, CutLER, and CLAID~(NCut), NCut~\cite{ncut} is adopted to carry out the partition. The NCut is computationally intensive as it involves the eigenvalue decomposition of a matrix of a large size. The time cost of one cut is much higher than that of the bisecting clustering adopted by CLAID. The processing time for CLAID~(NCut) is relatively shorter than the other methods based on NCut. The cut time becomes shorter since the cut is applied to the small subsets of the hierarchy. Owing to the efficiency of bisecting clustering, CLAID is 6$\times$ faster than CLAID~(Ncut).

To verify the ability of our detector in detecting the latent instance regions in an arbitrary image, we further study the performance of CLAID on \textit{COCO 20k} segmentation validation dataset. In this evaluation, CLAID is treated as an unsupervised instance detection method. The exhaustiveness is measured by the recall rate of the object instances in different categories that have been annotated in the dataset. The recall is evaluated with respect to IoU score. Six detectors Deep Spectral~\cite{deepspectral}, TokenCut~\cite{tokencut}, CutLER~\cite{cutclr}, U2Seg\cite{cvpr24:u2seg}, Orth\cite{cvpr24:sun} and VoteCut~\cite{cvpr24:cuvler} are considered in the comparison. The detection performance is shown in Figure~\ref{fig:exhaust}. As seen from the figure, CLAID outperforms these detectors by a large margin. On the one hand, it largely owes to the hierarchical decomposition without the depth constraint. It allows the exhaustive detection of the latent instances. On the other hand, the introduction of ``dummy node'' filters out the detected invalid instance regions.

\begin{table}
    \caption{Localization accuracy (mIoU) on \textit{Instance-160}, \textit{Instance-335}, and \textit{INSTRE}.}
	\centering
	\fontsize{8}{7}\selectfont 
	\begin{tabular}{lccc}
		\toprule
		Methods & Instance-160 & Instance-335 & INSTRE \\
		\cmidrule(lr){1-4}
		DASR~\cite{dasr} & 0.252 & 0.299 & 0.417 \\
		DASR*~\cite{dasr} & 0.260 & 0.286 & 0.369 \\
		CutLER~\cite{cutclr} & 0.262 & 0.488 & 0.709 \\
		U2Seg~\cite{cvpr24:u2seg} & 0.222 & 0.386 & 0.597 \\
		VoteCut~\cite{cvpr24:cuvler} & \textbf{0.401} & \textbf{0.607} & \textbf{0.777} \\
		CLAID & 0.377 & 0.527 & 0.696 \\
		\bottomrule
	\end{tabular}
	\label{tab:iou}
\end{table}

\subsection{Performance on Image Retrieval}
Generally speaking, image retrieval can be viewed as a multi-instance search task. Our method is, therefore, also feasible for conventional image retrieval. In this experiment, we study the effectiveness of CLAID on image retrieval. Following the same processing pipeline, a bunch of CLAID features are extracted from one candidate image. Only one feature is extracted from a query image. The ranking list is produced based on the position where a candidate's image first appears in the instance search ranking list.

Performance is evaluated on three benchmarks \textit{Paris6k}~\cite{paris}, \textit{Holidays}~\cite{holidays}, and \textit{Oxford5k}~\cite{oxford}. As shown from the Table~\ref{tab:imageres}, CLAID consistently outperforms a majority of representative methods in the literature. The best performance is achieved by the DUODIS~\cite{pami23:zhang}. DUODIS excels by mining latent objects directly from the test datasets and employs a pre-trained model, which is pre-conditioned on images of landmarks and buildings for feature extraction.

\begin{table}
\caption{The comparison on the processing time costs}
	\centering
	\fontsize{8}{7}\selectfont  
	\begin{tabular}{lrrc}
		\toprule
		Methods & $\#$cut & TM per image (s) & TM per cut (s)\\
		\toprule
		Deep Spectral~\cite{deepspectral} & 5 & 17.3 & 3.457\\
		TokenCut~\cite{tokencut} & 1 &9.6 & 9.600\\
		CutLER~\cite{cutclr} & 3 & 28.7 & 9.566\\
		VoteCut~\cite{cvpr24:cuvler} & - & 4.4 & 0.518\\
		CLAID~(NCut) & 29 & 14.6 & 0.503\\
		CLAID & 29 &\textbf{2.4} & \textbf{0.083}\\
		\bottomrule
	\end{tabular}
	\label{tab:speed}
\end{table}

\begin{table}
    \caption{Performance (mAP) on image retrieval. $^\ddag$ digits are from the referred paper.  $^{\dag}$ digits are from ~\cite{dasr}}
    \centering
    \fontsize{8}{7}\selectfont  
    \begin{tabular}{llll}
        \toprule
        Methods & Holidays & Oxford5k & Paris6k\\
        \toprule
         R-MAC~\cite{r-mac}$^\ddag$& - & 0.669 & 0.830\\
        CAM-weight~\cite{class-weighted} & 0.785$^{\dag}$ & 0.712$^\ddag$ & 0.805$^\ddag$  \\
        BLCF~\cite{blcf}  & 0.854$^{\dag}$ & 0.722$^\ddag$ & 0.798 $^\ddag$ \\
        BLCF-SalGAN~\cite{blcf}  & 0.835$^{\dag}$ & 0.746$^\ddag$ & 0.812$^\ddag$ \\
         DASR*~\cite{dasr}$^\ddag$& 0.873 & 0.613 & 0.744\\
         DUODIS~\cite{pami23:zhang}$^\ddag$& - &\textbf{0.900} &\textbf{0.952} \\
        CLAID & \textbf{0.899} &0.780 & 0.872\\
        \bottomrule
    \end{tabular}
    \label{tab:imageres}
\end{table}

\subsection{Ablation Studies}
\begin{table}
    \caption{Ablation study on the seed selection and the filtering strategy on \textit{INSTRE}}
	\centering
	\fontsize{8}{7}\selectfont 
	\begin{tabular}{lccccr}
		\toprule
		\multirow{2}{*}{Init.} &\multirow{2}{*}{\cancel{dummy}}&
		\multicolumn{4}{c}{INSTRE} \\
		\cmidrule(lr){3-6}
		& & mAP-50 & mAP-100 & mAP-all & \#Features  \\
		\midrule
		seed & $\times$ & 0.893 & 0.826 & 0.774 & 1,180,746\\
		rand & $\surd$ & \textbf{0.894} & \textbf{0.827} & \textbf{0.775} & 962,154 \\
		seed & $\surd$ & \textbf{0.894} & \textbf{0.827} & \textbf{0.775} & \textbf{827,426}  \\
		\bottomrule
	\end{tabular}
	\label{tab:ablation}
\end{table}

\begin{table}
\begin{center}
\caption{Instance Search Performance (mAP) when setting the decomposition depth as the termination threshold}
\label{tab:k_dep}
\begin{tabular}{lcrccc} 
\toprule
Datasets & Depth & \#Features & mAP-50 & mAP-100 & mAP-all \\\hline
\multirow{4}{*}{Instance-335} & 3 & 572,796 & 0.705 & 0.624 & 0.651 \\
							 & 5 & 2,536,668 & 0.727 & 0.647 & 0.670 \\
							 & 7 & 10,392,156 & 0.737 & 0.657 & 0.678 \\ \hline
	CLAID					 & - & 1,473,165 & \textbf{0.778} & \textbf{0.710} & \textbf{0.725} \\ \hline\hline
\multirow{4}{*}{INSTRE} & 3 & 383,102 & 0.857 & 0.768 & 0.686 \\
					   & 5 & 1,692,166 & 0.872 & 0.788 & 0.713 \\
					   & 7 & 6,932,422 & 0.881 & 0.801 & 0.729 \\ \hline    
	CLAID    	       & - & 827,426 & \textbf{0.894} & \textbf{0.827} & \textbf{0.775} \\ 
 	\bottomrule
\end{tabular}
\end{center}
\end{table}

To study the impact of each major component in CLAID, an ablation analysis is carried out on \textit{INSTRE}. Namely, we study the impact of seed selection (Section~\ref{sec:seed}) and the contribution from the filtering scheme on the ``dummy node''  (Section~\ref{sec:dummy}). Table~\ref{tab:ablation} shows the mAP results and the total number of features extracted from \textit{INSTRE} images under each configuration. As shown in the Table, compared to CLAID with random initialization, the detector produces \textit{16\%} less redundant instance regions. Moreover, compared to the run extracting features for dummy nodes, the standard configuration of CLAID shows slightly higher mAP while using  \textit{42\%} fewer features. This basically indicates that regions that correspond to the dummy nodes make no contribution to the instance-level feature representation.


In order to study the efficacy of the proposed termination condition (defined in Eqn.~\ref{eq:connect}), a variant run of the method is pulled out. In this study, the termination condition is replaced with the decomposition depth. The maximum decomposition depth is varied from \textit{3} to \textit{7}. The filtering with ``dummy node'' condition is disabled as well. The experiments are carried out on INSTRE and Instance-335. The instance search performance is compared to the standard configuration of CLAID and presented in Table~\ref{tab:k_dep}. The instance search  performance from  fixed-depth steadily increases when a larger decomposition depth is applied. Nevertheless, the higher performance is at the cost of an exponentially increasing number of regions, which is around eight times more than that of our adaptive decomposition. The large amount of redundant decomposition leads to a lot of invalid regions, which become the noises for the instance search. This is the main reason that our method still shows consistently better performance than the best performance from the fixed-depth decomposition. In particular, our method is $\textit{4}{\sim}\textit{5}$ percentage points higher in mAP-all.

\subsection{Configuration Studies}
\begin{figure}[t]
    \begin{center}
    \subfloat[]
    {\includegraphics[width=0.40\linewidth]{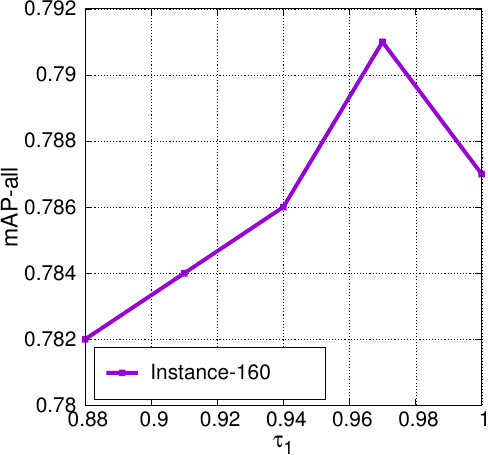}}
     \hspace{0.15in}
     \subfloat[]
    {\includegraphics[width=0.46\linewidth]{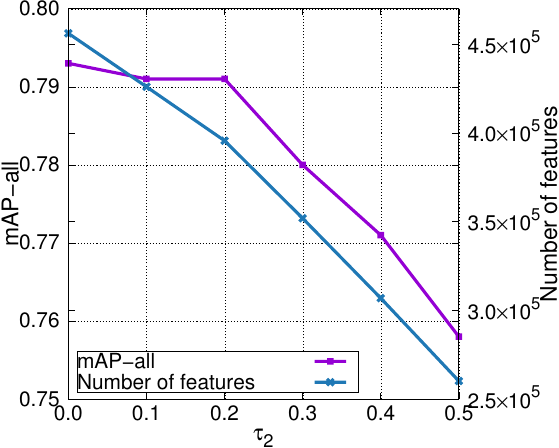}}
    \end{center}
    \caption{Experimental results of mAP with varying $\tau_1$ and $\tau_2$ settings on \textit{Instance-160}.}
    \label{fig:para}
\end{figure}

In our method, there are two key hyperparameters, namely the average internal connectivity ($\tau_1$) and the level of semantic dominance ($\tau_2$). $\tau_2$ is the threshold that controls whether a region is a ``dummy node''. The higher this threshold is, the more likely a region is to be filtered out as a ``dummy node''. To evaluate the impact of hyperparameter settings on our model's performance, the sensitivity tests have been pulled out. The experiments are carried out on \textit{Instance-160}. In the experiments, when validating $\tau_2$, we fix $\tau_1$ to \textit{0.97}. Conversely, when validating $\tau_1$, we fixed $\tau_2$ to \textit{0.2}. This setup allows us to isolate the impact of each parameter and analyze its individual impact on the model's performance. The instance search performance trends on \textit{Instance-160} are presented in Figure~\ref{fig:para}. As shown from the figure, the search performance is more sensitive to the level of semantic dominance ($\tau_2$), while only minor performance variation is observed when $\tau_1$ varies. Increasing $\tau_2$ progressively reduces the number of generated features. The search performance remains stable as long as $\tau_2$ is in the range $[0.0,~0.2]$. Nevertheless, at $\tau_2 = 0.5$, more than one-third of the generated features are filtered out. At the same time, the mAP remains close to 0.76, indicating that the ``dummy node'' mechanism effectively removes redundant features while maintaining good retrieval performance.

\section{Conclusions}
We have presented a novel instance-level feature, namely CLAID. A novel top-down bisecting procedure has been introduced to decompose the feature set, which is produced by a pretrained ViT, recursively into a hierarchy. To support the decomposition, an adaptive termination condition and the ``dummy node'' filtering strategy have been proposed. Each node on the hierarchy corresponds to a potential instance region in the image. As a consequence, we are able to decompose an image into a collection of instance regions in different granularities and represent them with features of uniform length. The features are class-agnostic because the detector is built on a non-classification ViT backbone. These features may cover a dominant image region, an integral of multiple instances, or a single instance. It, therefore, supports the similarity measurement between two images as well as two instances. As shown by the comprehensive experiments, excellent performance is achieved on instance search as well as image retrieval. Additional experiments also show the high compatibility of our method with different pretrained ViT backbones, such as DINOv2 and SigLIP. In particular, the excellent performance on SigLIP does indicate the feasibility of our method to unify text-to-instance search, instance search, and image retrieval into one search framework.

\section*{Acknowledgment}
This work is supported by Ningbo Municipal Bureau of Science and Technology‌ 2024Z121 and Natural Science Foundation of Fujian Province  2025J01002, 2024J01904.

\bibliographystyle{IEEEtran}
\bibliography{qysun}
\begin{IEEEbiography}[{\includegraphics[width=1in,height=1.25in,clip,keepaspectratio]{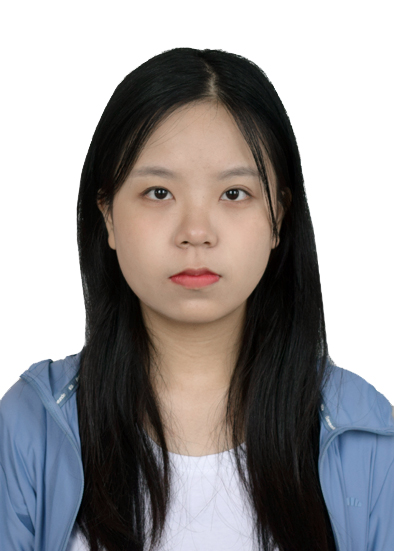}}]
{Qi-Ying Sun} received her bachelor degree from Soochow University in 2022. She is currently a graduate student in Xiamen University. Her research interest is about the visual instance search.
\end{IEEEbiography}
\begin{IEEEbiography}[{\includegraphics[width=1in,height=1.25in,clip,keepaspectratio]{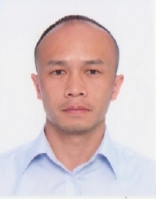}}]
{Wan-Lei Zhao} received his Ph.D degree from City University of Hong Kong in 2010. He received M.Eng. and B.Eng. degrees in Department of Computer Science and Engineering from Yunnan University in 2006 and 2002 respectively. He is a professor with School of Information Science at Xiamen University, China. Before joining Xiamen University, he was a Postdoctoral Scholar in INRIA, France. His research interests include multimedia information retrieval and video processing.
\end{IEEEbiography}
\begin{IEEEbiography}[{\includegraphics[width=1in,height=1.25in,clip,keepaspectratio]{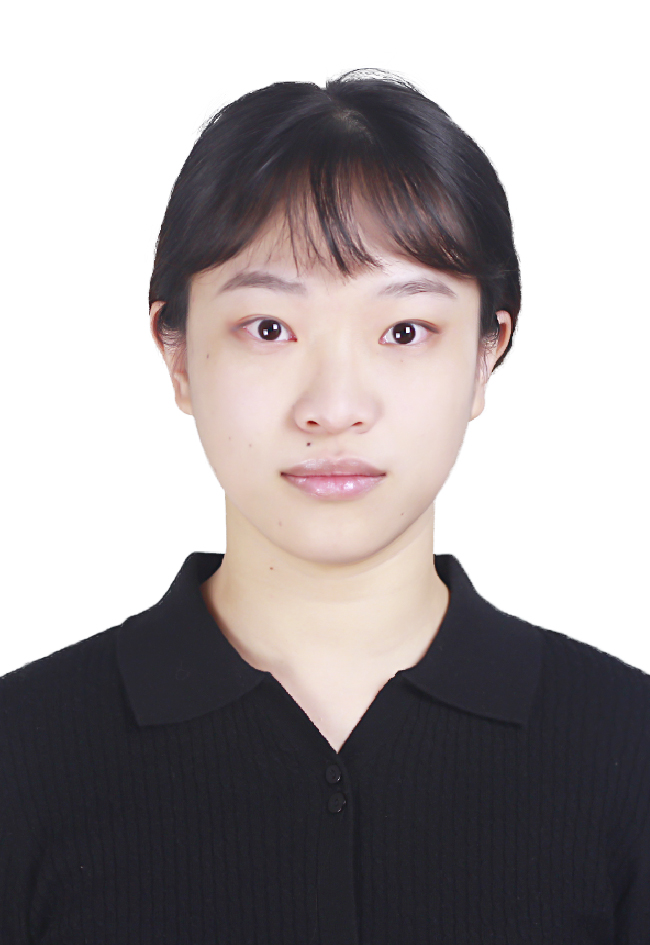}}]
{Hui-Ying Xie} received her Master degree from Tianjin University in 2021. She is currently a Ph.D student in Xiamen University. Her research interest is about the visual instance search.
\end{IEEEbiography}
\begin{IEEEbiography}[{\includegraphics[width=1in,height=1.25in,clip,keepaspectratio]{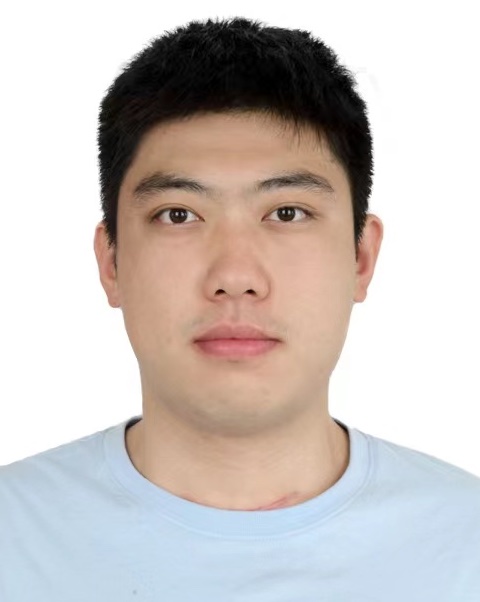}}]
{Yi-Bo Miao} received his bachelor degree from Xiamen University in 2021. He is currently a graduate student in Xiamen University, China. His research focus is multimedia information retrieval.
\end{IEEEbiography}
\begin{IEEEbiography}[{\includegraphics[width=0.9in,clip,keepaspectratio]{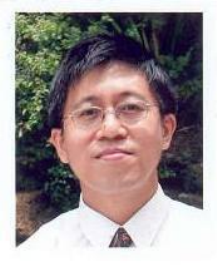}}]
{Chong-Wah Ngo} received the B.Sc. and M.Sc. degrees both in computer engineering from Nanyang Technological University of Singapore, Singapore, and the Ph.D. degree in computer science from Hong Kong University of Science and Technology, Hong Kong. He is a Professor with School of Computing and Information Systems, Singapore Management University. Before joining Singapore Management University, he worked with the Department of Computer Science, City University of Hong Kong, Hong Kong. He was a Postdoctoral Scholar with the Beckman Institute, University of Illinois, Urbana-Champaign. His main research interests include large-scale multimedia information retrieval, video computing, multimedia mining and visualization. He was an Associate Editor for the IEEE TRANSACTIONS ON MULTIMEDIA and is currently steering committee member of TRECVid, ICMR (International Conference on Multimedia Retrieval) and ACM Multimedia Asia. He is program Co-Chair of ACM Multimedia 2019, and general Co-Chairs of ICIMCS 2018 and PCM 2018. He was named ACM Distinguished Scientist in 2016 for contributions to video search and semantic understanding.
\end{IEEEbiography}

\newpage
\appendix
\begin{figure*}[t]
	\centering
	\includegraphics[width=0.95\textwidth]{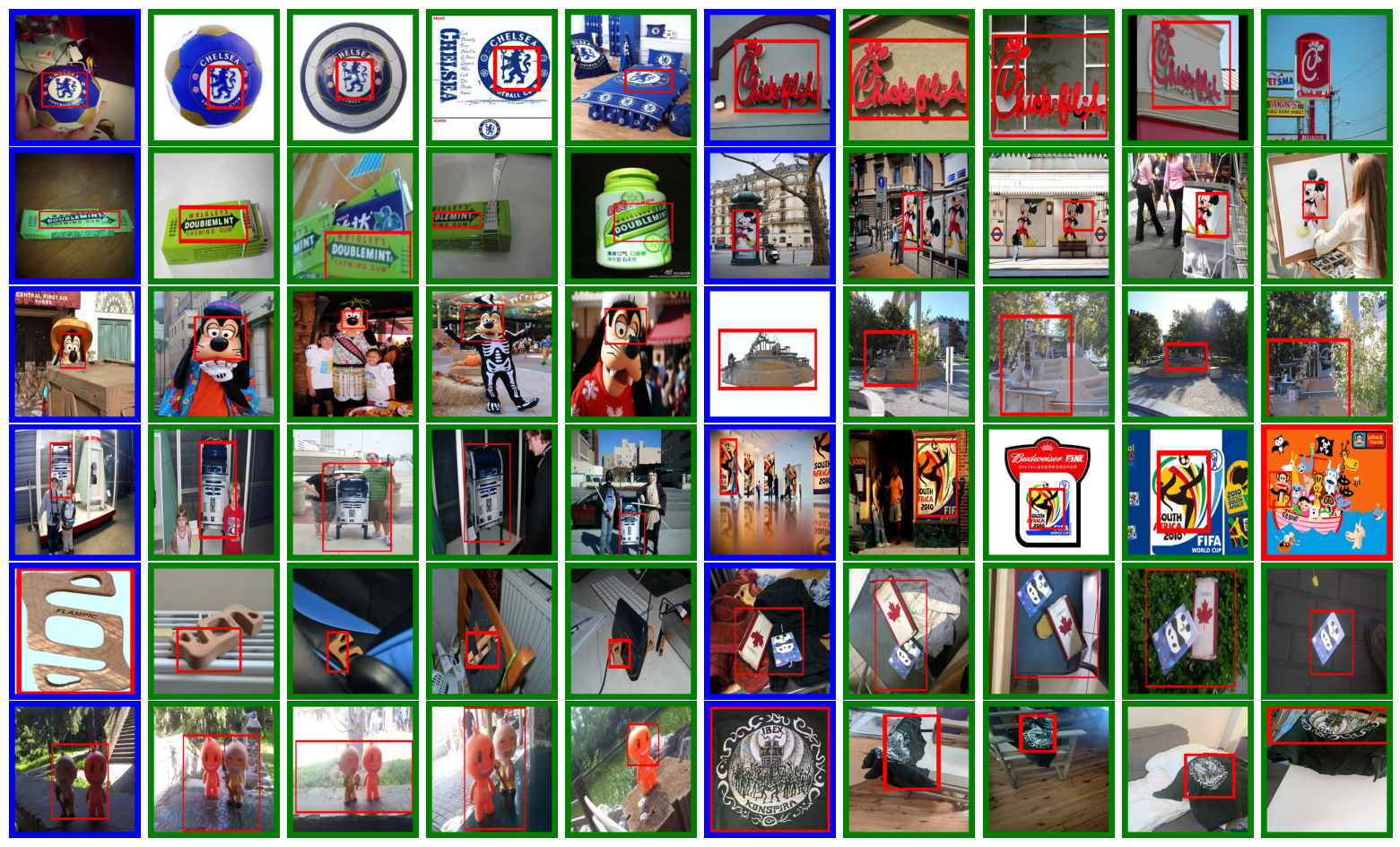}
	\caption{More visual results of instance search. The examples are selected from INSTRE and ILIAS.}
	\label{fig:moreres}
\end{figure*}

\textbf{More Retrieval Result Showcases} Figure~\ref{fig:moreres} showcases more results of instance search by CLAID. The instance queries are from images enclosed in a blue border. Each query is followed by four retrieved images from the top-1, top-5, top-10, and top-50 outcomes for that query. True-positives are colored in green, while false-positives are in red. Our method performs well on the majority of queries, including instances with deformation (left column Row-3, Row-4, Row-5; right column Row-3), occlusion (left column Row-4; right column Row-2), and other complex scenarios. It also shows good results in multi-instance problems. This confirms the robustness of the method we proposed.

\textbf{The Search Performance with Different Detector Backbone} Theoretically speaking, a detector can be built upon any ViT networks, e.g., DINO~\cite{DINO}, DINOv2~\cite{dinov2}, SigLIP~\cite{Zhai2023SigLIP}, and ViT pretrained by~\cite{vit}. Table~\ref{tab:other} shows the search results based on different ViT backbones. The experiments are conducted on \textit{INSTRE}. As seen from the table, relatively poor performance is achieved with ``ViT-B/16 (pretrained)'' because this backbone is a trained classification model, which is not class-agnostic. The best performance is achieved with the language-image model SigLIP. 

\textbf{Multi-modality Instance Search Performance} Supported SigLIP backbone, our method is further evaluated on the dataset mini-ILIAS~\cite{ilias2025}, which is a large-scale dataset designed to evaluate the performance of multi-modality instance search. There are \textit{1,232} queries and \textit{5} million images in the reference set. Both the text-to-instance search (T2I) and instance-to-instance search (I2I)\footnote{including multi-instance and single-instance search.} have been evaluated. The best performance reported in~[51] is treated as the comparison baseline. This method is also based on SigLIP. The search performance is presented in Table~\ref{tab:miniilias}. As shown in Table~\ref{tab:miniilias}, with the same backbone as the baseline method, CLAID shows significantly better performance on both tasks.  This does show the feasibility of CLAID to unify the text-to-instance search, image search, and instance search into one search framework.

\begin{table}
    \caption{Performance (mAP) on INSTRE utilizing different detector backbone}
	\centering
		\fontsize{8}{7}\selectfont 
	\begin{tabular}{lccc}
		\toprule
		Backbones & mAP-50&mAP-100&mAP-all\\
		\midrule
		ViT-B/8~(DINO)&0.894 &0.827 &0.775 \\
          ViT-B/16~(DINO) & 0.896&0.828& 0.774  \\
          ViT-S/8~(DINO) & 0.886&0.817& 0.762  \\
          ViT-B/14~(DINOv2) & 0.897 & 0.829 & 0.776  \\
          ViT-B/16~(SigLIP) & \textbf{0.993}& \textbf{0.979}& \textbf{0.959} \\
          ViT-B/16~(pretrained) & 0.861&0.778& 0.696  \\
		\bottomrule
	\end{tabular}
	\label{tab:other}
\end{table}

\begin{table}
    \centering
    \caption{Multi-modality Instance Search (mAP) on mini-ILIAS. $^\ddag$ digits are from the referred paper}
    \begin{tabular}{llccc}
        \toprule
        Task & Methods & mAP-50 & mAP-100 & mAP-1K \\
        \midrule
        \multirow{2}{*}{I2I} 
            & SigLIP~\cite{ilias2025}$^\ddag$ & - & - & 0.275 \\
            & CLAID (SigLIP) & 0.582 & 0.584 & \textbf{0.585} \\
        \midrule
        \multirow{2}{*}{T2I} 
            & SigLIP~\cite{ilias2025}$^\ddag$ & - & - & 0.142 \\
            & CLAID (SigLIP) & 0.394 & 0.398 & \textbf{0.402} \\
        \bottomrule
    \end{tabular}%
    \label{tab:miniilias}
\end{table}

\textbf{Further Evaluation on ILIAS Core} To investigate the effects of backbone choice and model capacity, CLAID is further evaluated on ILIAS Core using several variants of SigLIP and SigLIP2. We evaluate different backbone configurations, where region detection and feature description use either identical or distinct backbones. The results for text-to-instance search (T2I) and instance-to-instance search (I2I) are reported in Table~\ref{tab:ilias_core_backbones}. CLAID is tested with backbones of varying capacities and can benefit from larger models. In particular, when the detector backbone is fixed, adopting a larger descriptor backbone improves search performance on both tasks. The best performance on both tasks is achieved by combining different detector and descriptor backbones. These results demonstrate the flexibility and effectiveness of CLAID's decoupled detection and description stages, allowing each stage to benefit from a suitable backbone independently.

\begin{table*}[!t]
    \caption{Multimodal instance search performance (mAP) on ILIAS Core
    with different detector and descriptor backbones.}
    \label{tab:ilias_core_backbones}
    \centering
    \fontsize{8.5}{10}\selectfont
    \setlength{\tabcolsep}{3pt}
    \renewcommand{\arraystretch}{1.15}
    \begin{tabular*}{\textwidth}{@{\extracolsep{\fill}}llcccccc@{}}
        \toprule
        \multirow{2}{*}{\shortstack[l]{Detector\\backbone}} &
        \multirow{2}{*}{\shortstack[l]{Descriptor\\backbone}} &
        \multicolumn{3}{c}{I2I} &
        \multicolumn{3}{c}{T2I} \\
        \cmidrule(lr){3-5}
        \cmidrule(lr){6-8}
        & &
        mAP@50 & mAP@100 & mAP@1000 &
        mAP@50 & mAP@100 & mAP@1000 \\
        \midrule
        SigLIP1 Base/16-512 &
        SigLIP1 Base/16-512 &
        0.839 & 0.841 & 0.841 &
        0.761 & 0.763 & 0.764 \\

        SigLIP1 SO400M/14-384 &
        SigLIP1 SO400M/14-384 &
        0.830 & 0.832 & 0.833 &
        0.782 & 0.783 & 0.784 \\

        SigLIP2 SO400M/16-512 &
        SigLIP2 SO400M/16-512 &
        0.856 & 0.858 & 0.858 &
        0.774 & 0.775 & 0.776 \\
        \midrule
        SigLIP1 Base/16-512 &
        SigLIP1 SO400M/14-384 &
        \textbf{0.879} & \textbf{0.880} & \textbf{0.880} &
        \textbf{0.819} & \textbf{0.820} & \textbf{0.820} \\
        \bottomrule
    \end{tabular*}
\end{table*}
\end{document}